\documentclass[twoside,11pt]{article}
\usepackage{blindtext}
\usepackage{jmlr2e}

\usepackage{type1cm} 
\usepackage{makeidx}       
\usepackage{graphicx}  
\usepackage{dsfont}
\usepackage{amsmath}
\usepackage{multicol} 
\usepackage{algorithmic}
\usepackage{algorithm}
\usepackage{float}
\DeclareMathOperator*{\argmax}{arg\,max}

\newcommand{\pscal}[2]{\left\langle #1, #2 \right\rangle}

\newcommand{\Abs}[1]{\left\lVert #1 \right\rVert}
\usepackage[bottom]{footmisc}
\usepackage{newtxtext}       %
\usepackage[varvw]{newtxmath}     
\usepackage{lastpage}
\jmlrheading{23}{2022}{1-\pageref{LastPage}}{1/21; Revised 5/22}{9/22}{21-0000}{ Equal contribution}
\ShortHeadings{Tackling Computational Heterogeneity in FL}{Tackling Computational Heterogeneity in FL}
\firstpageno{1}
\begin{document}
\title{Tackling Computational Heterogeneity in FL:\\ A Few Theoretical Insights}
\author{\name Adnan Ben Mansour 
       \email adnan.mansour@be-ys-research.com \\
       \addr  be-ys Research\\
       Argonay, 74370, France\\
       \AND
       \name Gaia Carenini \email gaia.carenini@ens.psl.eu \\
       \addr ENS - PSL University\\
             Paris, 75005, France\\
       \AND 
       \name Alexandre Duplessis \email alexandre.duplessis@ens.psl.eu \\
       \addr ENS - PSL University\\
             Paris, 75005, France\\}

\maketitle

\begin{abstract}
The future of machine learning lies in moving data collection along with training to the edge.
Federated Learning, for short FL, has been recently proposed to achieve this goal. The principle of this approach is to aggregate models learned over a large number of distributed clients, i.e., resource-constrained mobile devices that collect data from their environment, to obtain a new more general model. The latter is subsequently redistributed to clients for further training. A key feature that distinguishes federated learning from data-center-based distributed training is the inherent heterogeneity. In this work, we introduce and analyse a novel aggregation framework that allows for formalizing and  tackling computational heterogeneity in federated optimization, in terms of both heterogeneous data and local updates. Proposed aggregation algorithms are extensively analyzed from a theoretical, and an experimental prospective.
\end{abstract}


\section{Introduction}
Until recently, machine learning models were extensively trained in centralized data center settings using powerful computing nodes, fast inter-node communication links, and large centrally-available training datasets. However, with the proliferation of mobile devices that collectively gather a massive amount of relevant data every day, centralization is not always practical [\cite{Edge(1)}]. Therefore, the future of machine learning lies in moving both data collection and model training to the edge to take advantage of the computational power available there, and to minimize the communication cost.
Furthermore, in many fields such as medical information processing, public policy, and the design of products or services, the collected datasets are \emph{privacy-sensitive}. This creates a need to reduce human exposure to data to avoid confidentiality violations due to human failure. This may preclude logging into a data center and performing training there using conventional approaches. In fact, conventional machine learning requires feeding training data into a learning algorithm and revealing information indirectly to the developers. When several data sources are involved, a merging procedure for creating a single dataset is also required, and merging in a privacy-preserving way is still an important open problem [\cite{Merging(1)}].

\begin{algorithm}[H]
\caption{General Federated Learning Protocol}
\label{alg:generic_FL}
\textbf{Input}: \emph{N}, \emph{C}, \emph{T}, \emph{E} \\
\textbf{Output}: $w_{TE}$
\begin{algorithmic}[1] 
\STATE Initialize $w_0$.
\FOR{each round $t\in\{0,E, 2E,\dots,(T-1)E\}$}
\STATE $m \leftarrow \max(C \cdot N, 1)$
\STATE $I_t\leftarrow$\textsc{Create-Client-Set}$(m)$ 
\FOR{each client $i \in I_t$ \textbf{in parallel}}
\STATE $w_{t+E}^{i} \leftarrow \text{\textsc{Client-Update}}(w_t)$
\ENDFOR
\STATE $w_{t+E} \leftarrow \text{\textsc{Aggregation}}(w_{t+E}^1, \dots, w_{t+E}^N)$
\ENDFOR
\STATE \textbf{return} $w_{TE}$
\end{algorithmic}
\end{algorithm}

Recently, \cite{FL-Intro(1)} proposed a distributed data-mining technique for edge devices called \emph{Federated Learning} (FL), which allows to decouple the model training from the need for direct access to the raw data. Formally, FL is a protocol that operates according to Algorithm \ref{alg:generic_FL}, cf. \cite{SurveyFL(1)} for an overview. The framework involves a group of devices called \emph{clients} and a \emph{server} that coordinates the learning process. Each client has a local training dataset that is never uploaded to the server. The goal is to train a global model by aggregating the results of the local training.
Parameters fixed by the centralized part of the global learning system include:  $N$ clients, the ratio of clients $C$ selected at each round, the set of clients $I_t$ selected at round $t$, the number of communication rounds $T$, and the number of local epochs $E$. A model for a client $i$ at a given instant $t$ is completely defined by its weights $w_i^t$. At the end of each epoch $t \in \{ 0, \dots, TE -1 \}$, $w_{t+1}^i$ indicates the weight of client $i \in I$. For each communication round $t \in \{ 0, E, \dots, (T-1)E \}$, $w_t$ is the global model detained by the server at time $t$, and $w_{TE}$ is the final weight. In the following, we will use the notations given in Table \ref{tab:notations}.\\

\begin{table}[!t]
\centering
\caption{Conventions used in this paper.\label{tab:notations}}      
\begin{tabular}{p{4cm}p{9.5cm}}
\hline\noalign{\smallskip}
\textbf{Notation} & \textbf{Meaning}\\
\hline
  $N$   &~~number of clients\\
  $C$  &~~ratio of clients\\
  $I_t$ &~~set of clients selected at round $t$\\
  $T$ &~~number of communication rounds~~\\
  $E$&~~number of local epochs\\
  $w_t$&~~weights of the global model at time $t$\\
  $w_t^i$&~ model of client $i$ at time $t$\\
  $F_i^t$&~~loss function of the $i$-th client at time $t$\\
  $F$&~~averaged loss function at time $t$\\
  $\eta_t$&~~learning rate decay at time $t$\\
  $\zeta_t^i$&~~mini-batch associated to client $i$ at time $t$\\
   $^*$&~~index of optimality\\
\hline
\end{tabular}
\end{table}    

Algorithm \ref{alg:generic_FL} describes the training procedure for FL. The framework involves a fixed set of $I = \{1,\dots,N\}$ clients, each with a local dataset. Before every communication round $t \in \{0, E, \dots, (T-1)E \}$, the server sends the current global model state to the clients and requests them to perform local computations based on the global state and their local dataset, and sends back an update. At the end of each round, the server updates the weights of the model by aggregating the clients' updates, and the process repeats.
For the client selection procedure (\textsc{Create-Client-Set}), local training procedure (\textsc{Client-Update}), and aggregation of the local updates (\textsc{Aggregation}), several possibilities exist. For some results concerning client selection, see [\cite{Aggregation(1), Aggregation(2), Aggregation(3), Aggregation(4)}]. Regarding local updates, available methods range from simple variants of SGD, such as mini-batch SGD [\cite{Learning(1)}], to more sophisticated approaches, such as PAGE [\cite{Learning(2)}]; other results are included in [\cite{Learning(3), Learning(4), Learning(5), Learning(6)}]. We will describe in greater detail the existing routines for aggregation, the central topic of this work.
In 2017, the seminal work of \cite{FL-Intro(1)} proposed a plain coordinate-wise mean averaging of model weights; later, \cite{FLNeural(1)} proposed an extension that takes the invariance of network weights under permutation into account. The same year, \cite{FLNeural(2)} proposed an auto-tuned communication-efficient secure aggregation. More recently, \cite{FLNeural(3)} extended the coordinate-wise mean averaging approach, substituting it with a term that amplifies the contribution of the most informative terms over less informative ones. Then, \cite{Pervasive(1)} adjusted this to enforce closeness of local and global updates.  Last year, \cite{FLNeural(4)} introduces an aggregation that allows clients to select what values of the global model are sent to them.
Despite methodological advances, there is neither theoretical nor practical evidence for the right criterion for choosing a particular aggregation strategy.
\begin{figure}
    \centering
    \includegraphics[scale=0.35]{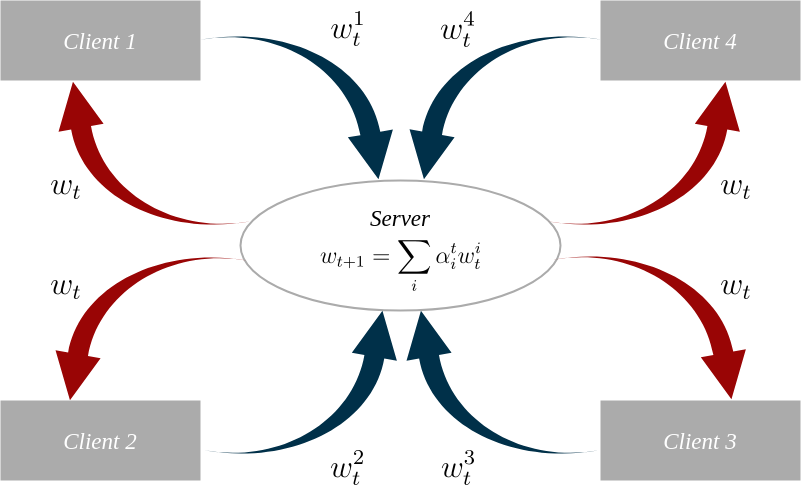}
    \caption{Simplified representation of classic FL framework.}
    \label{fig:my_label}
\end{figure}

\subsection{The Challenges from Computational Heterogeneity in FL}
As we have seen above, several emerging FL algorithms have been proposed. Due to the high cost of real deployment, existing studies in FL usually involves simulations [\cite{Sim(1),Sim(2),Sim(3)}] and have no data to describe how devices participate in FL [\cite{Data(1)}]. The direct consequence of this approach is that this studies build on excessively ideal assumptions, for instance the one that all the devices are constantly available for training and equipped with the same resources, e.g., the same CPU and RAM capacity [\cite{Sim(1),Sim(2),Sim(3), Agno(1), Agno(2)}]. However, these assumptions can be inadequate for FL deployment in practice.  FL, in fact, requires a large
number of devices to collaboratively accomplish a learning task, which poses a great challenge, namely \emph{heterogeneity} [\cite{Het(1)}], that impacts FL both in terms of accuracy and
training time. We can divide heterogeneity in two main macro-classes: the \emph{system heterogeneity} and the \emph{statistical heterogeneity}.\\

In federated settings, system heterogeneity points out the significant variability in the systems characteristics across the network, as
devices may differ in terms of hardware, network connectivity, and battery power. These systems characteristics make issues such as stragglers significantly more prevalent than in typical
data center environments. Several solution to handle systems heterogeneity have been proposed, e.g. asynchronous communication, see [\cite{Com(1), Com(2)}], active device sampling, see [\cite{Com(3)}], and fault tolerance, see [\cite{Fault(1)}]. Statistical heterogeneity deals instead with the challenges that arise when training federated models from data that is not identically distributed across devices,
both in terms of modeling the data, and in terms of analyzing the convergence
behavior of associated training procedures. There exists a large body of literature in machine learning that has modeled statistical heterogeneity
via methods such as meta-learning and multi-task learning [\cite{Meta(1),Meta(2),Meta(3),Meta(4)}].\\

Despite heterogeneity is associated with several possible problems such as \emph{free-riding} [\cite{Free-ride(1)}], theoretical guarantee to convergence of heterogeneous federated learning  have been recently found [\cite{Guar(1), Guar(2), Guar(3)}] and approaches to overcome these challenges formalized, e.g. thanks to the introduction of  \emph{Personalized Federated Learning} (PFL) [\cite{PFL(1),PFL(2),PFL(3)}], and the one of heterogeneous ensemble knowledge transfer [\cite{Het(2)}]. Several methods have been proposed to attack the heterogeneity arising from specific sources such as data, see [\cite{Het(3), Het(4), Het(5)}], partial and biased client participation, see [\cite{Het(6), Aggregation(4)}]. In what follows, we will discuss how to possibly tackle heterogeneous local updates performances in edge clients, propose new aggregation methods, test them experimentally, and provide insights on their convergence properties, their stability and the client participation within the training.

\section{Tackling Performance-Heterogeneity in FL: The Theoretical Side}
We study theoretically how the heterogeneous performances of clients can be exploited in aggregation methods (under reasonable assumptions). The analysis presented is fairly general and  allows to extract information concerning the existing trade-off between accuracy and efficacy. This analysis can be seen as a remarkable follow-up of \cite{FLNeural(3)}, the first work presenting a convergence
analysis of federated learning with biased client selection that is cognizant of the training progress
at each client, and the work where was discovered that biasing the client selection towards clients with higher local losses increases the rate of convergence (compared to unbiased client selection).

\subsection{Framework of Analysis \& Preliminaries} \label{framework}
\noindent Throughout the analysis, we assume that all the clients are involved in each local and global iteration, i.e., $C=1$. We denote with $F_i$ the loss function of the $i$-th client, and with $F$ the weighted average of the $F_i$ upon the distribution $P:=\{p_i\mid i\in I\}$. We restrain our analysis to the case in which the \textsc{Client-Update} procedure is mini-batch SGD with learning rate decay $\eta_t$ and mini-batches $\zeta_t^i$ of cardinality $b$. In particular: 
\begin{equation}
    g_i(w_t^i):=\frac{1}{b}\sum\limits_{\zeta\in \zeta_t^i} \nabla F_i(w_t^{i},\zeta)
\end{equation}
In any iteration, the weights of the model are defined as follows:
 \begin{equation}
   w_{t+1}^i := \left\{
    \begin{array}{ll}
        \displaystyle w_t^i - \eta_t g_i(w_t^i) & \mbox{if } E\hspace{-0.5em}\not{|}t \\
        \\
        \displaystyle\sum_{j\in I} \alpha_t^j (w_t^j - \eta_t g_j(w_t^j)) := w_{t+1} & \mbox{if } E | t
    \end{array} \right.
\end{equation}
\noindent where $\alpha_t^j$ is the \emph{aggregation coefficient} referred to client $j$ at communication round $t$, and where, for each $t$, the following constraint holds: 
\begin{equation}
   \sum\limits_{j \in I}{\alpha_t^j} = 1 
\end{equation}
In our mathematical analysis, we introduce a few assumptions:\vspace{0.15cm}\\
\textbf{Assumption 1} (\textit{$L$-smoothness}) $F_1,\dots,F_N$ satisfy:
\begin{center}
$\forall v,w$, $F_{i}(v)\leq F_i(w)+\pscal{v-w}{\nabla F_i(w)}+\frac{L}{2}\Abs{v-w}_2^2$   
\end{center}
\textbf{Assumption 2} (\textit{$\mu$-convexity})
$F_1,\dots,F_N$ satisfy: 
\begin{center}
$\forall v,w$, $F_{i}(v)\geq F_i(w)+\pscal{ v-w}{\nabla F_i(w)} +\frac{\mu}{2}\Abs{v-w}_2^2$    
\end{center}
\textbf{Assumption 3} The variance of the stochastic gradient descent is bounded, more formally, the following condition is satisfied:
\begin{center}
$\forall i\in I$, $\mathds{E}\Abs{g_i(w_i)-\nabla F_i(w_i)}^2\leq \sigma^2$  
\end{center}
\textbf{Assumption 4} The stochastic gradient's expected squared norm is uniformly bounded, in mathematical terms: 
\begin{center}
$\forall i\in I$, $\mathds{E}\Abs{g_i(w_i)}^2\leq G^2$  
\end{center} 
What follows is closely related to what was  previously done in \cite{Aggregation(4)}, the novelty arise from the fact that: (a) instead of analyzing the selection of clients, we examine the attribution of the weights to them, and (b) we extensively study the expression of the \emph{learning error} from which we derive principled aggregation strategies.\vspace{0.2cm}\\
\noindent To facilitate the convergence analysis, we define the quantity $w_t$ (for which $t\neq 0 \bmod E$) as:
  \begin{equation}
    w_{t+1}:=w_t - \eta_t\sum\limits_{i\in I} \alpha_t^i g_i(w_t^i)
\end{equation}  
\noindent where $\alpha_t^i=p_i$.
Let $w^\star$ be the global optimum of $F$ and $w^\star_i$ the global optimum of $F_i$. We  define $F^\star$ as $F(w^\star)$, $F^\star_i$ as $F(w^\star_i)$ and  \emph{heterogeneity} as:
 \begin{equation}
    \Gamma:=F^\star-\sum_{i\in I} p_i F_i^\star
\end{equation}   
\noindent We list below a couple of results useful in proving the main theorem.
\begin{lemma}\label{smoothRes}
Let $f$ be a $L$-smooth function with a unique global minimum at $w^\star$. Then :
\begin{equation}
    \forall w,\hspace{1cm}||\nabla f(w)||^2\leq 2L(f(w)-f(w^\star))
\end{equation}    
\end{lemma}
\begin{lemma}\label{discrepency}
With the same notations as above and defining $\mathds{E}[.]$ as the total expectation over all random sources, the expected average discrepancy between $w_t$ and $w_t^i$ is bounded:
 \begin{equation}
   \displaystyle\mathds{E} \left[\sum_{i\in I} \alpha_t^i \Abs{w_t-w_t^i}^2\right] \le 16\eta_t^2 E^2 G^2
\end{equation}   
\end{lemma}
\noindent Before presenting the main results, we define the \emph{weighting skew} $\rho$\footnote{We observe that $\rho(t,w)$ is not defined when $F(w)=\sum\limits_{i\in I} p_i F_i^\star$. This condition will be always assumed below.} as:
\begin{equation}\label{rho}
    \rho(t,w):= \frac{\sum\limits_{i\in I}\alpha_t^i (F_i(w)-F_i^\star)}{F(w)-\sum\limits_{i\in I} p_i F_i^\star}
\end{equation}    
and introduce these notations: 
$\overline{\rho} := \min\limits_{t = 0 \hspace{-0.5em} \mod E} ~\rho(t,w_t)$, and $\Tilde{\rho} := \underset{t = 0 \hspace{-0.5em} \mod E}{\max}~ \rho(t,w^\star)$.
\subsection{Main Theorem and Consequences}
\noindent In the framework outlined, we state an extension of the main theorem presented in \cite{Aggregation(4)}, that is adapted to our extended goal. The proofs are available in the appendix \ref{app:theorem}.
\begin{theorem}\label{MainNONIID}
Under assumptions (1 - 4), the following holds:
 \begin{flalign}\label{main}
   \mathds{E}\left[\Abs{w_{t+1}-w^\star}^2\right] \le \left(1-\eta_t \mu\left(1+\frac{3}{8}\overline{\rho}\right)\right) \mathds{E}\left[\Abs{w_{t}-w^\star}^2\right]& \notag \\
   + \eta_t^2 \left(32 E^2 G^2 + 6 \overline{\rho} L \Gamma + \sigma^2\right) &  \\
   +2 \eta_t \Gamma \left(\Tilde{\rho}-\overline{\rho}\right)& \notag
\end{flalign}    
\end{theorem}
From Theorem \ref{MainNONIID}, we can directly deduce Corollary \ref{eq} below.
\begin{corollary}\label{eq}
Assuming $\eta_t = \frac{1}{\mu (t+\gamma)}$ and $\gamma = \frac{4L}{\mu}$, the following bound holds:
 \begin{equation}
\mathds{E}[F(w_T)] - F^\star \le \frac{1}{T+\gamma} \mathcal{V}(\overline{\rho}, \Tilde{\rho}) + \mathcal{E}(\overline{\rho}, \Tilde{\rho}) 
\end{equation}   
where:
\begin{align}
& \mathcal{V}(\overline{\rho}, \Tilde{\rho}) = \frac{4L(32\tau^2 G^2 + \sigma^2)}{3\mu^2 \overline{\rho}} + \frac{8L^2 \Gamma}{\mu^2} + \frac{L\gamma \Abs{w^0-w^\star}^2}{2} \notag\\
& \mathcal{E}(\overline{\rho}, \Tilde{\rho}) = \frac{8L\Gamma}{3\mu}\left(\frac{\Tilde{\rho}}{\overline{\rho}}-1\right) \notag\\ \notag
\end{align}   
\end{corollary}
\begin{remark}
Corollary \ref{eq} implies that: 
 \begin{equation}
    \mathbb{E}[F(w_T)-F^\star] = O(1/T)
\end{equation}    
\end{remark}  
The mathematical expressions $\mathcal{V}$  and $\mathcal{E}$ are estimators for the \emph{speed of convergence}, and the \emph{learning error}, respectively. A complex multi-objective optimization problem arises when trying to maximize the speed while minimizing the error. We decouple these two quantities and optimize them separately without underestimating the existing trade-off among them. This procedure allows to outline the global trends, but it does not imply the universal optimality of the strategies defined below. 
\begin{remark}
Since $\frac{8L^2 \Gamma}{\mu^2} + \frac{L\gamma \Abs{w^0-w^\star}^2}{2}$ is a constant depending only on the data and the initial guess, and  $\overline{\rho}$ may be arbitrary large, we can deduce from Corollary \ref{eq} the existence of a minimal value for the convergence speed, given by:
    \begin{equation}
        \mathcal{V}_{\min} := \frac{8L^2 \Gamma}{\mu^2} + \frac{L\gamma \Abs{w^0-w^\star}^2}{2}
    \end{equation}
\end{remark}
 In this framework, we can analyze all the possible scenarios, starting from the one in which $\Gamma = 0$, that can be appointed as \emph{error-free case} and corresponds to an IID-dataset.
\subsection*{Error-free Framework}
 Under the assumption that $\Gamma = 0$, the main theorem can be leveraged as follows:
 \begin{flalign}
      \mathds{E}\left[\Abs{w_{t+1}-w^\star}^2\right]  \le \left(1-\eta_t \mu\left(1+\frac{3}{8}\overline{\rho}\right)\right) \mathds{E}\left[\Abs{w_{t}-w^\star}^2\right]
      + \eta_t^2 \left(32 E^2 G^2 + \sigma^2\right)
\end{flalign}   
and applying Corollary \ref{eq}, we derive the following inequality:
\begin{equation}
    \mathds{E}[F(w_T)]-F^\star \le \frac{1}{T+\gamma}\left[\frac{4L(32\tau^2 G^2 + \sigma^2)}{3\mu^2 \overline{\rho}} + \frac{L\gamma \Abs{w^0-w^\star}^2}{2}\right]
\end{equation}  
Despite its simplicity, this setting is interesting since the error term vanishes, and therefore we can deduce a truly optimal algorithm given by the maximization of $\overline{\rho}$: \footnote{$\overline{\rho}$ is well defined as long as $F(w_t)\neq F(w^\star)$ for all $t$, which is a reasonable assumption.}, achieved when:
\begin{equation}
        \alpha_t^i = \left\{
    \begin{array}{ll}
        \displaystyle \frac{1}{|J_t|} & \mbox{if } i\in J_t \\
        \\
        \displaystyle 0 & \mbox{else } 
    \end{array}
\right.
\end{equation}
where $J_t= \underset{i\in I}{\argmax} (F_i(w_t)-F_i^\star)$.
\subsection*{General Framework}
 In the general case,  both $\mathcal{V}$ and $\mathcal{E}$ depend on the choice of the $\alpha_{i}^t$. As already noticed before, this raises a multi-objective problem that doesn't allow for a joint optimization of terms $\mathcal{V}$ and $\mathcal{E}$. Consequently, we provide an approximated optimization that builds upon the existing trade-off between the convergence speed and the accuracy\footnote{
It is important to notice that the bounds for $\mathcal{V}$ and $\mathcal{E}$ are not tight. Consequently we cannot guarantee the unconditional optimality of the  strategies proposed.}.
\begin{remark}
We observe that optimizing the convergence speed, while "forgetting" about the error, amounts to maximize $\overline{\rho}$, exactly as done in the error-free case. Instead, minimizing $\mathcal{E}(\overline{\rho}, \Tilde{\rho})$ neglecting $\mathcal{V}$, amounts to minimize $\frac{\Tilde{\rho}}{\overline{\rho}}-1$. This is achieved when $\alpha_t^i = p_i$, which gives $\mathcal{E}=0$.  
\end{remark}
 Now, knowing that $\alpha_t^i=p_i$ ensures obtaining optimal accuracy, we assume $\alpha_t^i=\kappa_t^i p_i$. The following notation is used:
 \begin{equation}
     \pi_t= \underset{i\in I}{\min}~\kappa_t^i, \Pi_t = \underset{i\in I}{\max~} \kappa_t^i, \pi = \underset{t}{\min}~\pi_t, \text{ and } \Pi = \underset{t}{\max~} \Pi^t
 \end{equation}
\begin{center}
 
\end{center}
Without loss of generality, we assume without that $\forall t, \pi_t > 0$. If it were not the case, we would have assigned to the $\alpha_t^i$ equal to zero an infinitesimal value, and increment the other  $\alpha_t^i$ substantially. Under these assumptions, we have that $\frac{\Tilde{\rho}}{\overline{\rho}} \le \frac{\Pi}{\pi}$, $\frac{1}{\overline{\rho}}\le \frac{1}{\pi}$ and therefore:
\begin{equation}
  \mathds{E}[F(w_T)]-F^\star \le \frac{1}{T+\gamma}\left[C + \frac{\lambda_1}{\pi} \right] + \lambda_2 \frac{\Pi - \pi}{\pi}  
\end{equation}      
where $C, \lambda_1$ and $\lambda_2$ are constants. Since,
 \begin{equation}
    \Pi \min~p_i \le \max~ \kappa_t^i p_i \le 1 -(N-1) \min~ \kappa_t^i p_i\le 1 - (N-1)\pi \min~p_i
\end{equation}   
we can infer that $\Pi \le \frac{1-(N-1)\pi \min~p_i}{\min~ p_i}$ and $\mathcal{E} \le \frac{1}{\pi \min~p_i}-N$, from which, we obtain:
\begin{equation}\label{Finalbound}
  \mathds{E}[F(w_T)]-F^\star \le \frac{1}{T+\gamma}\left[C + \frac{\lambda_1}{\pi} \right] + \lambda_2 \left(\frac{1}{\pi \min~p_i}-N\right)
\end{equation}   
\begin{remark}
 This last inequality has an intrinsic interest; in fact, it allows to state that the new speed and error bounds depend exclusively on $\pi$ and to ensure a bound on the error term (once set a properly chosen minimal value of the $\alpha_t^i$).    
\end{remark}
\subsection{Derived Aggregation Strategies}\label{strategy}
The theoretical results discussed above provides several important insights for the design of aggregation algorithms.\\

The first algorithm presented is the \textsc{generalized FedAvg}, that corresponds to take $\alpha_t^i = p_i$ for any $t$ and $i\in I$.
This strategy is inspired by \cite{FL-Intro(1)} and it boils down to consider the weighted average (upon $p_i$) of the local models as global model. As observed above, this approach is optimal in terms of accuracy (since $\mathcal{E}=0$) and its convergence speed can be bounded as below: 
\begin{equation}
 \mathcal{V} = \mathcal{V}_{\min} + \frac{4L-32\tau^2 G^2 + \sigma^2}{3\mu^2}
\end{equation}
The second algorithm proposed is called \textsc{FedMax} and it is defined as follows. For any $t$:
\begin{equation}
      \alpha_t^i = \left\{
    \begin{array}{ll}
        \displaystyle \frac{1}{|J_t|} & \mbox{if } i\in J_t \\
        \\
        \displaystyle 0 & \mbox{else } 
    \end{array}
\right.  
\end{equation}
where: 
\begin{equation}
    J_t= \underset{i\in I}{\argmax} (F_i(w_t)-F_i^\star)
\end{equation}
Note that two distinct clients in practice never have the same value, i.e. $|J_t|=1$. This strategy is our original algorithmic contribution, and consists in considering as global model the client's local model with the worst performance at the end of the previous communication round. This approach partially leverages the difference among the values of the loss functions of the different clients and, as observed above, this strategy gives an optimal bound on the convergence speed. For improving the performance in real-world applications and for avoiding over-training on outliers, we introduce a couple of variants of the previous algorithm, namely \textsc{FedMax($k$)} and \textsc{FedSoftMax}.\\

\textsc{FedMax($k$)}, instead of taking the client with the highest loss, considers the first $k$ clients when sorted by decreasing order with respect to $F_i(w_t)-F_i^\star$. This strategy boils down to the client selection strategy \emph{Power-of-Choice}, introduced in \cite{Aggregation(4)}. In \textsc{FedSoftMax}, 
for any $t$ and $i\in I$, we take $\alpha_t^i = p_i \exp (T^{-1} (F_i(w_t) - F^\star_i))$ re-normalized, i.e., the softened version of the original routine. The reason behind the introduction of this method is reinforcing the stability of FedMax, but this has, as well, the theoretical advantage of ensuring nonzero values of the $\alpha_i^t$. Note that, for this method, we can obtain an upper bound over the error by applying inequality \ref{Finalbound}.
\section{Tackling Performance-Heterogeneity in FL: The Practical Side}
One of the greatest difficulties in developing new algorithms in ML is to combine theoretical guarantees with practical requirements. With the aim of providing algorithms suitable for exploitation in applications, we conduct an experimental analysis with a twofold purpose to establish the performance of the proposed strategies and to identify their potential weaknesses and strengths. 
\subsection{Experimental Framework}
We describe below the full experimental framework involved in the study of the strategies described above. The design of the experimental apparatus is minimal; in fact, the goal is to focus maximally on the effects of the aggregation procedure.
\paragraph{Synthetic Data} We generate two distinct synthetic datasets, corresponding to the IID and to the non-IID framework. For the first, we  sort data according to labels, choose the cardinality of the different local datasets and distribute the items preserving an identical label distribution over the clients. Instead, for the second, we sort the dataset by label and divide it into multiple contiguous shards according to the criteria given in \cite{FL-Intro(1)}. Then, we distribute these shards among our clients generating an unbalanced partition. We do not require clients to have the same number of samples, however each client has at least one shard. We actually also implemented a "hand-pick" splitting system, to enable better control of the distribution of numbers among clients. Both methods were tested and gave similar results for all experiments.

\paragraph{Model} The model\footnote{Much better performance could be achieved using more complex models developed throughout literature. In this work, the performance of the network on the task is secondary and therefore we opt for the simplest model used in practice.} used is fairly basic: a CNN with two $3\times 3$ convolution layers (the first with 32 channels, the second with 64, each followed with $2\times 2$ max pooling), a fully connected layer with 1600 units and ReLu activation, and a final softmax output layer. The local learning algorithm is mini-batch SGD with a batch size fixed at $64$.

\paragraph{Parameters} The parameters involved in the experimental analysis are summarized in Table \ref{tab:parameters}. 

 \paragraph{Tasks description}
The task used is a classification task of the images of the datasets [MNIST \cite{MNIST}], and [Fashion-MNIST \cite{F-MNIST}], both in IID and in not-IID framework.    
\paragraph{Evaluation}
To evaluate the performance of the strategies proposed, we focus our attention on two kinds of measures: the \emph{accuracy} reached after a fixed number of communication rounds and the  index $R_{90}$, that corresponds to the number of communication rounds required to reach a $90\%$ accuracy. We furthermore keep track of the accuracy value and of the loss function at each communication round of the global model. 

\paragraph{Resources}
All the strategies are implemented within Pytorch and trained on an Intel Xeon E5-2670 2,60 Ghz,, 8 hearts, 64 Go RAM.

 \begin{table}[!t]
\caption{Parameters used in the experiments.\label{tab:parameters}}      
\begin{tabular}{p{2cm}p{7.5cm}p{3.5cm}}
\hline\noalign{\smallskip}
\textbf{Symbol} & \textbf{Meaning} & \textbf{Value}\\
\hline
  $N$   &~~number of clients & 50\\
  $C$  &~~ratio of clients & 1\\
  $N$   &~~size of each client's dataset in non-IID fr. & 200\\
  $\#_{\text{shard}}$&~~cardinality of the shards in non-IID fr. & 60\\
    $\#_{\text{shard, v}}$&~~cardinality of the shards in very-non-IID fr. & 100\\
  $T$ &~~number of communication rounds~~ & 50\\
  $E$&~~number of local epochs & 2\\
  $\eta_t$&~~learning rate at time $t$ & $10^{-4}\cdot 0.99^r$\\
  $b$&~~cardinality of the batch & 64\\
\noalign{\smallskip}\hline\noalign{\smallskip}
\end{tabular}
\end{table}
\subsection{Experimental Analysis}\label{experimental}
\begin{figure}[H]
\centering
\includegraphics[width=\linewidth]{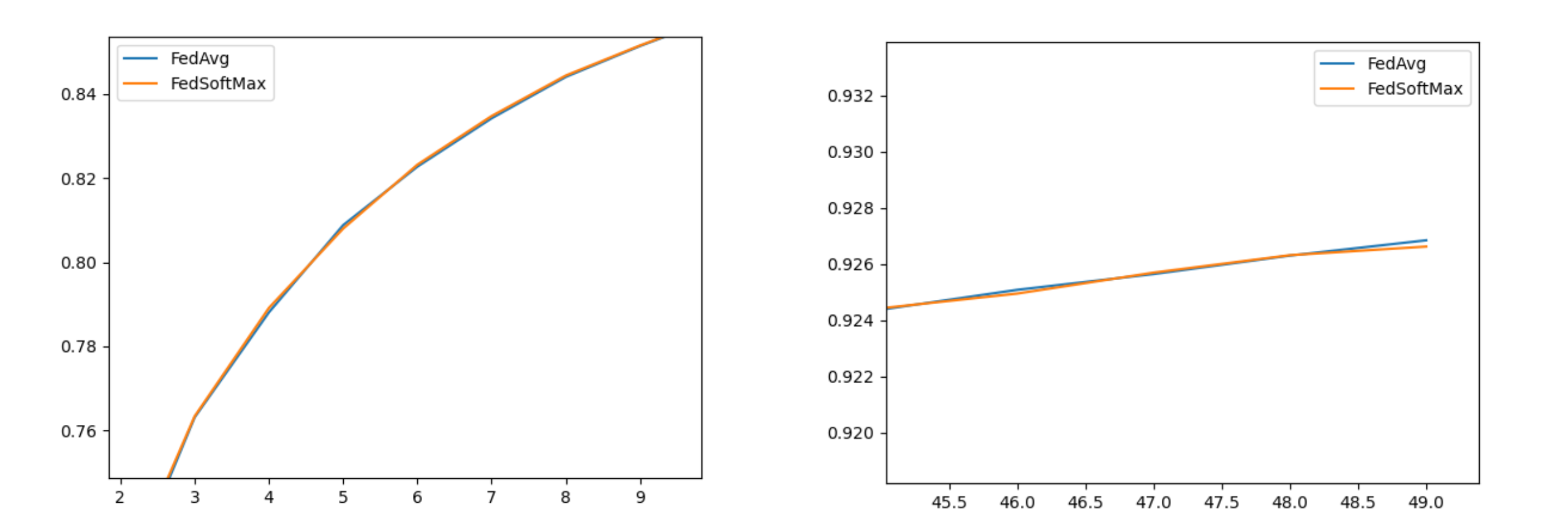}
\caption{\textbf{Comparative analysis between FedAvg and FedSoftMax: Final and intermediate accuracy in IID framwork.}\\The horizontal axis accounts for communication round and the vertical axes accounts for the accuracy reached. These results are the ones obtained on MNIST.}
\label{fig:Delta}       
\end{figure}

\begin{figure}
\centering
\includegraphics[width=\linewidth]{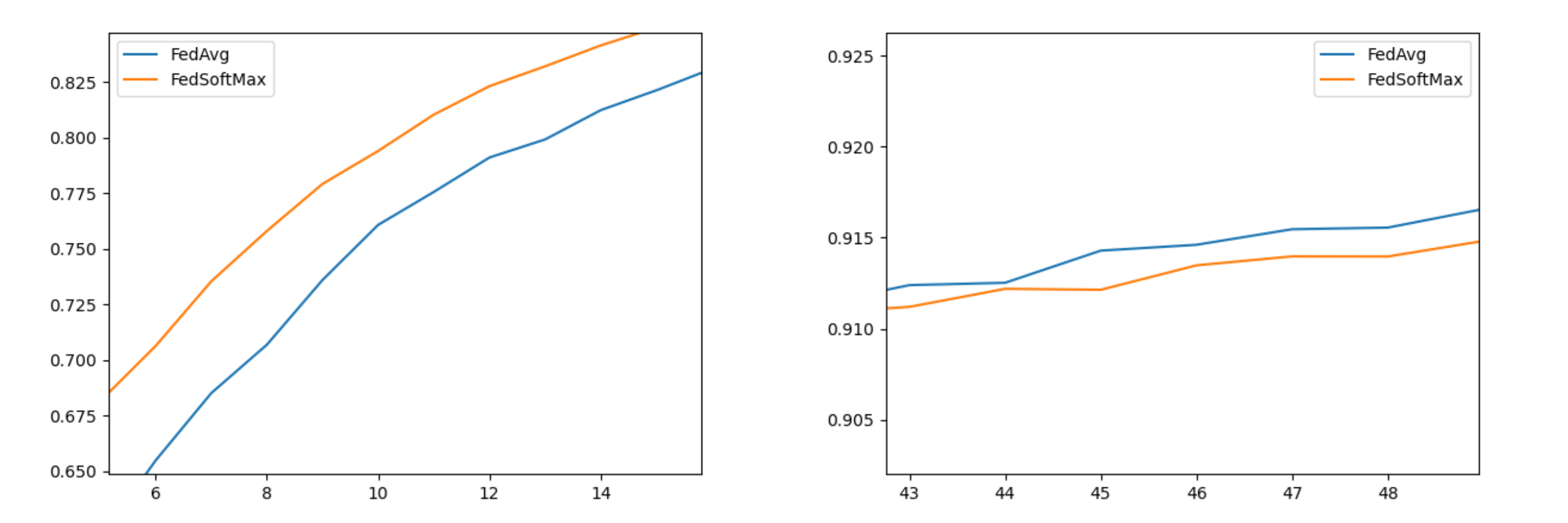}
\caption{\textbf{Comparative analysis between FedAvg and FedSoftMax: Final and intermediate accuracy in not IID framework.}\\The horizontal axis accounts for communication round and the vertical axes accounts for the accuracy reached. These results are the ones obtained on MNIST.}
\label{fig:Delta2}     
\end{figure}

\begin{figure}
\centering
\includegraphics[width=\linewidth]{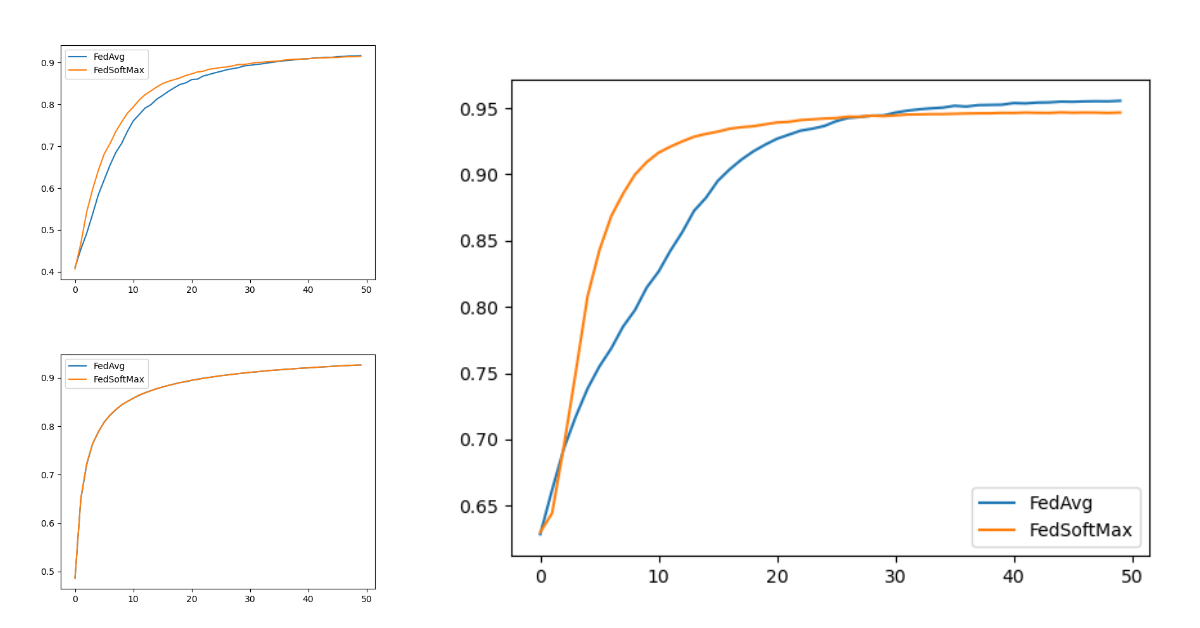}
\caption{\textbf{Comparative analysis between FedAvg and FedSoftMax.}\\The horizontal axis accounts for communication round and the vertical axes accounts for the accuracy reached. The image  bottom left refers to the IID framework, the top left to the not IID framework and the right one to the extremely not-IID framework. These results are the ones obtained on MNIST.}
\label{fig:Accuracy}       
\end{figure}

\begin{figure}
\includegraphics[width=\linewidth]{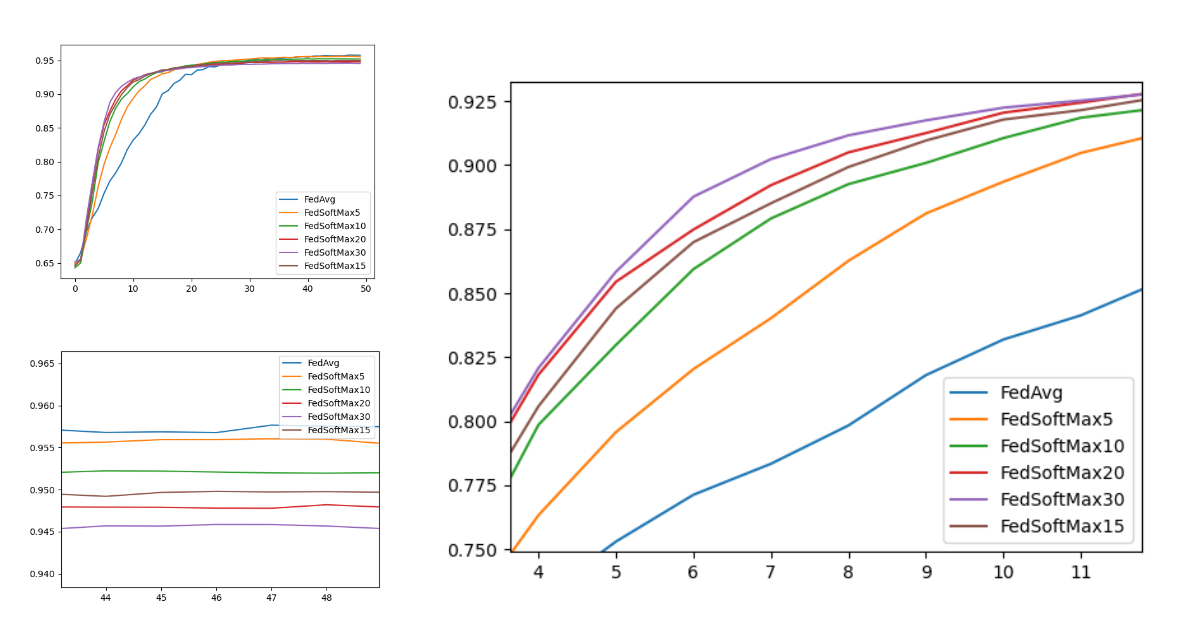}
\caption{\textbf{Comparative analysis between FedAvg and FedSoftMax(T) for several values of T, the temperature.}\\The horizontal axis accounts for communication round and the vertical axes accounts for the accuracy reached. These results are the ones obtained on MNIST in not IID framework.}
\label{fig:Max}       
\end{figure}

We focus on results related to FedMax, the main method introduced.
\paragraph{Comparative Analysis of the Strategies Proposed}
We have tested extensively the methods proposed in IID, not-IID and extremely not IID framework. We can observe that, in last two cases, it is sufficient to focus on the first 50 communication rounds in order to encounter a significant discrepancy among the methods.
While the difference between the final accuracy obtained through FedSoftMax and FedAvg is rather low in any framework (see Figures \ref{fig:Delta} and \ref{fig:Delta2}), a large gap is evident in how quickly the learning system achieves $90\%$ accuracy in the not-IID and very-not-IID cases (TNIID) (see Figure \ref{fig:Accuracy} and see Table \ref{tab:R}). Therefore, experiments give a clear confirmation of the theory, and tend to prove that the upper bound provided by the main theorem is quite tight. 
\begin{remark}
 FedSoftMax has a higher convergence speed compared to FedAvg. The discrepancy increments with the bias of the data with respect to the closest IID distribution.
   Moreover, FedSoftMax produces a rather small bias that is directly proportional to the distance that the data distribution among the clients has from the IID one.    
\end{remark}
To try to better understand the optimality of FedSoftMax, we have investigated the change in the performance when it is modified the parameter $T$, the one accounting for the temperature. The experimental results show that, if we restrict the temperatures considered to the range between 5 and 30, an higher temperature entails  an higher convergence speed (see Figure \ref{fig:Max}).

\begin{figure}[H]
\centering
\includegraphics[width=\linewidth]{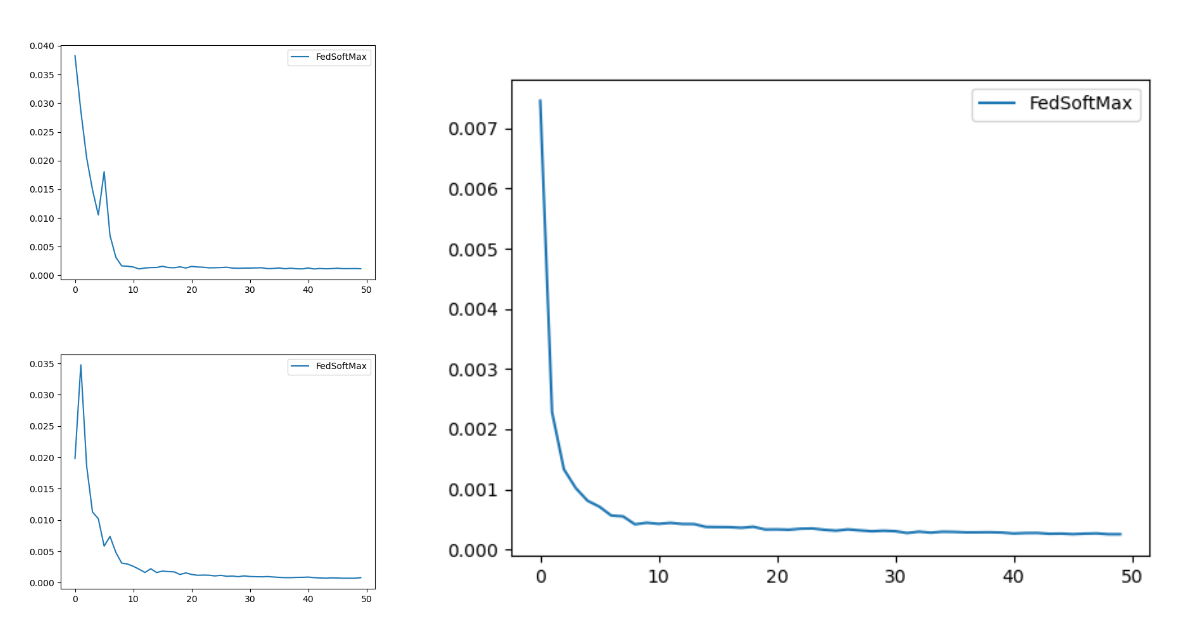}
\caption{\textbf{Convergence of $\alpha_i$ to $p_i$ as a function of time (represented through the communication rounds) in the IID framework.}\\The horizontal axis accounts for communication round and the vertical axes accounts for the difference among $\alpha_i$ and $p_i$. These results are the ones obtained on MNIST.}
\label{fig:alpha}       
\end{figure}

 \begin{table}
\caption{Results concerning convergence speed ($R_{90}$).\label{tab:R}}      
\begin{tabular}{p{2cm}p{3.5cm}p{8.5cm}}
\hline\noalign{\smallskip}
Framework & Algorithm & Confidence Interval for $R_{90}$\\
\hline
  TNIID   &  FedAvg & 14.877887807248147 - 15.842112192751852 \\
  TNIID  & FedSoftMax & 7.855914695373678 - 8.704085304626322\\
  NIID  & FedAvg & 12.726276322973177 - 13.957934203342614 \\
  NIID & FedSoftMax & 9.996249956446318 - 11.31953951723789\\
  IID & FedAvg & 21.527442685457743 - 24.58366842565337\\
  IID & FedSoftMax & 21.607110478485172 - 24.504000632625942\\
\noalign{\smallskip}\hline\noalign{\smallskip}
\end{tabular}
\end{table}
\newpage
\paragraph{Weakness and Strength of the Strategies Proposed}
One potential weakness that has emerged from the theory is that the method potentially converges to a different value of the optimum. We were therefore interested in studying whether experimentally a significant difference could be observed and whether this might preclude the use of the method. With this purpose, we studied the evolution of $\alpha_i$. The result is extremely positive, not only do we see that the imbalance produced is minimal, but rather we observe that the $\alpha_i$ almost always converges to the $p_i$ with a rate of $1/t$ (see Figure \ref{fig:alpha}). All these entails the following remark: 
\begin{remark}
  FedSoftMax is natural smooth interpolation between FedAvg and FedMax($k$), taking the advantage from the higher convergence speed of FedMax($k$) in the initial phase and the stability and correctness of FedAvg in the rest of the learning.   
\end{remark}

\begin{figure}[H]
\includegraphics[width=\linewidth]{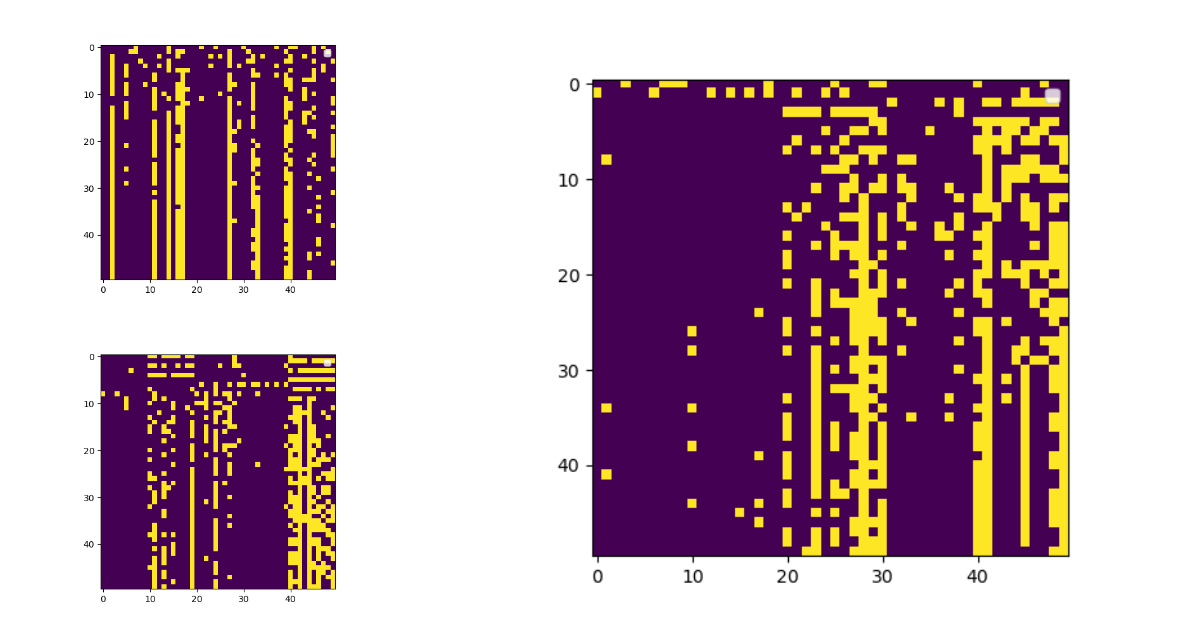}
\caption{\textbf{Clients participation in the aggregation process}\\The horizontal axis accounts for a unique identifier of the client and the vertical axis accounts for the communication round. The value of the $\alpha_i$ is encoded through colors. Then ten highest values are colored in yellow, while the others are colored in blue. These results are the one obtained on MNIST.}
\label{fig:part}       
\end{figure}

In fact FedMax(k) methods, while speeding the process at the beginning, give poor results when it comes to the final accuracy. This can actually be well visualized by analyzing the top losses of clients during FedSoftMax running (see Figure \ref{fig:part}). We actually see that only a small group of clients are used through all the process, and while this is profitable for speed purposes in the first rounds, it has a huge drawback in the following rounds since we only use a small amount of data that (by non-IIDness) is not representative of the whole dataset. There lies the power of FedSoftMax, which enables to use both the speed-up ability of FedMax, and the data of all clients at the end as in FedAvg.

Finally, we became interested in measuring the stability of FedSoftMax when compared to FedAvg. For this purpose we use a lag one autocorrelation measure based on the normalized standard deviation of the variations over one round. The results in this case show a more pronounced tendency toward instability than FedSoftMax, which nevertheless appears to be reasonably stable (see Table \ref{tab: stab}).
 \begin{table}[H]
 \centering
\caption{Results concerning stability.\label{tab: stab}}      
\begin{tabular}{p{2cm}p{3.5cm}p{5.5cm}}
\hline\noalign{\smallskip}
Framework & Algorithm & Smoothness value\\
\hline
  TNIID   &  FedAvg &  1.3686297084899968\\
  TNIID  & FedSoftMax & 2.322656907880106\\
  NIID  & FedAvg &  1.3404421487057407\\
  NIID & FedSoftMax & 1.7307167213190393\\
  IID & FedAvg &  2.8107170960158876\\
  IID & FedSoftMax & 2.7724249879251603\\
\noalign{\smallskip}\hline\noalign{\smallskip}
\end{tabular}
\end{table}
\subsection{Discussion \& Final Remarks}
We have extended the insightful analysis already carried out by \cite{Aggregation(4)}, and examined further the joint evolution of $\overline{\rho}$ and $\Tilde{\rho}$, obtaining simpler bounds. Taking advantage of these theoretical insights, we  have proposed a family of aggregation strategies, among which FedSoftMax is the most relevant one. Here, we complement our previous work by investigating empirically the latter with the goal of finding out  weaknesses and of quantifying its strength for potential exploitation in practice. The experimental results fully confirm the theory and also suggest that the bias introduced by mismatched weighting of the data distribution does not affect the quality of the final results. Moreover, this method seems to naturally converge to FedAvg leveraging the biases introduced in the first communication rounds. 
\subsection*{Further directions of research}
Several aspects emerged may be object of further analysis. We report them as associated research questions. From a theoretical point of view, we propose 3 possible directions to investigate:\vspace{0.2cm}\\ 
\noindent\textbf{Proposed Research Question 1}
Is it possible to obtain expressive bounds weakening at least one of the 4 assumptions introduced? We believe interesting results could be obtained by weakening of assumption 3.\\ 

\noindent\textbf{Proposed Research Question 2}
Can we substitute the learning algorithm used throughout all the analysis, i.e. SGD mini-batch, with others? We believe that interesting results may come even using fairly natural algorithms such as GD.\\

\noindent\textbf{Proposed Research Question 3}
The experiments have shown that the $\alpha_i$ coefficients converge to the $p_i$ in a framework where datasets are not too much non-IID. We might thus be interested in both proving this affirmation under supplementary hypothesis, and to see its consequences when it comes to the adaptation of the main theorem.

\noindent Moreover, we actually see that in the case of a very non-IID dataset the $\alpha_i$ actually do not converge to the $p_i$, but they still converge to some fixed limits, and it would be interested to study these limits, and their potential correlation with the $F_i^*$ or other client dependent parameters.\\

\noindent\textbf{Proposed Research Question 4}
Figure \ref{fig:Max} showed the correlation between parameter $T$ of the FedSoftMax method and the gain in speed. Further experiments not shown here show that increasing further $T^{-1}$ increases the speed up to a certain limits, i.e. the accuracy curves tend to converge to a "maximal-speed" curve. Not only could we empirically study the properties of this limit curve, but we could also try to give theoretical evidence for this observation.\\

\noindent\textbf{Proposed Research Question 5}
Can we extend our results to the non-convex setting? We suggest to start introducing some simplifying conditions such as 
the ones associated to the Polyak-Łojasiewicz inequality.    
\\
From a practical point of view, it could be interesting to investigate if there is a practical advantage induced by the speed-up given by FedSoftMax.

\subsection*{Acknowledgements}
\emph{This work was granted access to HPC resources of MesoPSL financed by the Region Ile de France and the project Equip@Meso (reference ANR-10-EQPX-29-01) of the programme \emph{Investissements d'Avenir} supervised by the Agence Nationale pour la Recherche.}

\newpage
\bibliography{sample}
\newpage
\appendix
\section{Proof of the Main Theorem \& its Corollary}
\textbf{Proof (Main Theorem)}
The first step of the proof consists in rewriting the argument of the expectation at the \texttt{LHS} of Inequality \ref{main}. 
In particular, we apply the definition of $w_{t+1}$, add and subtract the same quantity, and eventually develop the square. This leads to the following expression.
 \begin{flalign*}
    \Abs{w_{t+1}-w^\star}^2
    &= \Abs{w_t - w^\star}^2 - 2 \eta_t \pscal{w_t - w^\star}{\sum\limits_{i\in I} \alpha_t^i \nabla F_i(w_t^i)} \\
    & + 2 \eta_t \pscal{w_t - w^\star - \eta_t \sum\limits_{i\in I} \alpha_t^i \nabla F_i(w_t^i)}{\sum\limits_{i\in I} \alpha_t^i \nabla F_i(w_t^i) - \sum\limits_{i\in I} \alpha_t^i g_i(w_t^i)}\\
    & + \Abs{\eta_t \sum\limits_{i\in I} \alpha_t^i \nabla F_i(w_t^i)}^2 + \eta_t^2 \Abs{\sum\limits_{i\in I} \alpha_t^i \nabla F_i(w_t^i) - \sum\limits_{i\in I} \alpha_t^i g_i(w_t^i)}^2
\end{flalign*}
For seek of simplicity, we denote the addends at the \texttt{RHS}, as follows: 
\begin{equation*}
 =\Abs{w_t-w^\star}^2 + A_1 +A_2 + A_3 + A_4
\end{equation*}
The rest of the proof consists in bounding, sequentially, each of these addends.\vspace{0.2cm}\\
Concerning term $A_1$, we apply Cauchy-Schwartz inequality and by exploiting the convexity of the functions involved. Then, we derive the consequences of Assumption 1 and take advantage from the fact that $\nabla F_i(w_i^\star)=0$. This leads to the following upper bound for $A_1$: 
\begin{flalign*}
\le \sum\limits_{i\in I} \alpha_t^i \Abs{w_t - w_t^i}^2 + 2L \eta_t^2 \sum\limits_{i\in I} \alpha_t^i (F_i(w_t^i)-F_i^\star) - 2 \eta_t \sum\limits_{i\in I} \alpha_t^i \pscal{w_t^i - w^\star}{\nabla F_i(w_t^i)}
\end{flalign*}
then, by using Assumption 2, we can rewrite the last term of the previous inequality and applying Lemma 1, we obtain the final upper bound for $A_1$. These steps are reported below:
\begin{flalign*}
A_1&\le - 2 \eta_t \sum\limits_{i\in I} \alpha_t^i (F_i(w_t^i)-F_i(w^\star) + \frac{\mu}{2} \Abs{w_t^i-w^\star}^2) \\
&\le \eta_t^2 E^2 G^2 - \eta_t \mu \sum\limits_{i\in I} \alpha_t^i\Abs{w_t^i-w^\star}^2 + 2L \eta_t^2 \sum\limits_{i\in I} \alpha_t^i (F_i(w_t^i)-F_i^\star) \\
&- 2 \eta_t \sum\limits_{i\in I} \alpha_t^i (F_i(w_t^i)-F_i(w^\star))
\end{flalign*}
For bounding term $A_2$, we observe that, for the unbiasedness of the gradient estimator, $\mathds{E}(A_2)=0$. The bound for $A_3$ requires exclusively the application of Assumption 1:
\begin{flalign*}
    A_3 &= \Abs{\eta_t \sum\limits_{i\in I} \alpha_t^i \nabla F_i(w_t^i)}^2
    \le 2L\eta_t^2 \sum\limits_{i\in I} \alpha_t^i (F_i(w_t^i)-F_i^\star)
\end{flalign*}
Then, we obtain the bound for $A_4$ by applying Jensen's inequality, exploiting the linearity of the expected value and using Assumption 3.
\begin{flalign*}
    A_4
    &\le \eta_t^2 \sum\limits_{i\in I} \alpha_t^i \sigma^2 \le \eta_t^2 \sigma^2
\end{flalign*}
This sequence of bounds allows us to write the following expression:
\begin{flalign*}
    \mathbb{E}[\Abs{w_{t+1}-w^\star}^2]
    \le &(1-\eta_t \mu) \mathbb{E}[\Abs{w_t-w^\star}^2] + 16\eta_t^2 E^2 G^2 + \eta_t^2 \sigma^2 &\\
    &\quad + 4 L \eta_t^2 \mathbb{E}[\sum\limits_{i\in I} \alpha_t^i (F_i(w_t^i)-F_i^\star)] &\\
    &- 2 \eta_t \mathbb{E}[\sum\limits_{i\in I} \alpha_t^i (F_i(w_t^i)-F_i(w^\star))]
\end{flalign*}
Renaming the latter terms, we have that:
\begin{flalign*}
    &= (1-\eta_t \mu) \mathbb{E}[\Abs{w_t-w^\star}^2] + 16\eta_t^2 E^2 G^2 + \eta_t^2 \sigma^2 + A_5
\end{flalign*}
Bounding $A_5$, it's slightly more complicated, but the sequence of operations required is fairly similar to the one done above. The final upper bound is the following:
\begin{flalign*}
    A_5 
    &\le \eta_t^2 (16 E^2 G^2 + 6 \overline{\rho} L \Gamma) - \frac{3}{8}\eta_t \mu \overline{\rho} \mathbb{E}[\Abs{w_t-w^\star}^2] + 2 \eta_t \Gamma (\Tilde{\rho}-\overline{\rho})
\end{flalign*}
The proof is completed as follows: 
\begin{flalign*}    
    \Abs{w_{t+1}-w^\star}^2 &\le (1-\eta_t \mu) \mathbb{E}[\Abs{w_t-w^\star}^2] + 16\eta_t^2 E^2 G^2 + \eta_t^2 \sigma^2 + A_5\\
    &\le (1-\eta_t \mu(1+\frac{3}{8}\overline{\rho})) \mathbb{E}[\Abs{w_t-w^\star}^2] + \eta_t^2 (32 E^2 G^2 + 6 \overline{\rho} L \Gamma + \sigma^2) \\
    &+ 2 \eta_t \Gamma (\Tilde{\rho}-\overline{\rho})
\end{flalign*}

\label{app:theorem}
\noindent \textbf{Proof (Corollary)}
The proof is fairly simple and brief.
We start rewriting the main Theorem, as follows:
\begin{flalign*}
    \Delta_{t+1} \le (1-\eta_t\mu B)\Delta_t + \eta_t^2 C + \eta_t D
\end{flalign*}
where: 
\begin{center}
  $B = (1+\frac{3}{8}\overline{\rho})$, $C = 32 E^2 G^2 + 6 \overline{\rho} L \Gamma + \sigma^2$, and $D = 2 \Gamma (\Tilde{\rho}-\overline{\rho})$   
\end{center}
Let $\psi$ be the $\max \left\{\gamma\Abs{w_0-w^\star}^{2}, \frac{1}{\beta \mu B-1}\left(\beta^{2} C+D \beta(t+\gamma)\right)\right\}$,
where $\beta>\frac{1}{\mu B}$, $\gamma>0$.\\
The proof proceed by induction; from this argument, we derive that:
\begin{equation*}
 \forall t, \Delta_{t} \leq \frac{\psi}{t+\gamma}   
\end{equation*}
Then, by the L-smoothness of $F$, we obtain the following upper bound that concludes the proof.
\begin{flalign*}
\mathbb{E}\left[F\left(w_t\right)\right]-F^\star \leq \frac{L}{2} \Delta_{t} \leq \frac{L}{2} \frac{\psi}{\gamma+t}
\end{flalign*}

\end{document}